\documentclass[11pt,a4paper]{article}
\usepackage[hyperref]{acl2019}
\usepackage{times}
\usepackage{latexsym}
\usepackage[utf8]{inputenc}

\usepackage{url}
\usepackage{breakurl}

\usepackage{graphicx}
\usepackage{caption}
\usepackage{subcaption}

\usepackage{multicol}
\usepackage{ctable}

\usepackage{amsmath}

\usepackage[nolist,nohyperlinks]{acronym}
\begin{acronym}
    \acro{nlp}[NLP]{Natural Language Processing}
    \acro{am}[AM]{Argumentation Mining}
    \acro{sota}[SotA]{State-of-the-Art}
    \acro{nmt}[NMT]{Neural Machine Translation}
    \acro{lm}[LM]{Language Model}
    \acro{lstm}[LSTM]{Long short-term memory}
    \acro{blstm}[Bi-LSTM]{bidirectional LSTM}

    \acro{baseline}[\(\text{baseline}\)]{Baseline}
    \acro{baseline-i}[\(\text{baseline}_{+input}\)]{Baseline input attention}
    \acro{baseline-e}[\(\text{baseline}_{+error}\)]{Baseline input attention}
    \acro{bilstm}[\(\text{bilstm}\)]{Single Bi-LSTM}
    \acro{bilstm-i}[\(\text{bilstm}_{+input}\)]{Single Bi-LSTM input attention}
\end{acronym}

\aclfinalcopy 

\title{Is It Worth the Attention?\\A Comparative Evaluation of Attention Layers for\\Argument Unit Segmentation}

\author{
	Maximilian Spliethöver \and
	Jonas Klaff \and
	Hendrik Heuer \\
	University of Bremen\\
	Bibliothekstraße 1, 28359 Bremen, Germany\\
	\texttt{\{mspl,joklaff,hheuer\}@uni-bremen.de}}

\date{}

\begin{document}
\maketitle

\begin{abstract}
    Attention mechanisms have seen some success for natural language processing downstream tasks in recent years and generated new State-of-the-Art results.
    A thorough evaluation of the attention mechanism for the task of Argumentation Mining is missing, though.
    With this paper, we report a comparative evaluation of attention layers in combination with a bidirectional long short-term memory network, which is the current state-of-the-art approach to the unit segmentation task.
    We also compare sentence-level contextualized word embeddings to pre-generated ones.
    Our findings suggest that for this task the additional attention layer does not improve upon a less complex approach.
    In most cases, the contextualized embeddings do also not show an improvement on the baseline score.
\end{abstract}

\section{Introduction}
\ac{am} is increasingly applied in different fields of research like fake-news detection \citep{cabrio_five_2018} and political argumentation and network analysis\footnote{See for example the MARDY project (\url{https://www.socium.uni-bremen.de/projekte/?proj=570&print=1}, last accessed: 2019-04-15, 09:50UTC+2).}.\\
One crucial part of the \ac{am} pipeline is to segment written text into argumentative and non-argumentative units.
Recent research in the area of unit segmentation \citep{eger_neural_2017,ajjour_unit_2017} has lead to promising results with F1-scores of up to $0.90$ for in-domain segmentation \citep{eger_neural_2017}.
Nevertheless, there is still a need for more robust approaches.\\
Given the recent progress of attention-based models in \ac{nmt} \citep{bahdanau_neural_2014,vaswani_attention_2017}, this paper evaluates the effectiveness of attention for the task of argumentative unit segmentation.
The idea of the attention layers added to the recurrent network is to enable the model to prioritize those parts of the input sequence that are important for the current prediction \citep{bahdanau_neural_2014}.
This can be achieved by learning additional parameters during the training of the model.
With the additional information gained, the model learns a better internal representation which improves performance.\\
Additionally, we evaluate the impact of contextualized distributed term representations (also referred to as word embeddings hereinafter) on all our models.
The goal of word embeddings is to represent a word as a high-dimensional vector that encodes its approximate meaning.
This vector will be generated by a model trained on a language modeling task, like next-word prediction \citep{mikolov_efficient_2013}, for a given text corpus.
The approximation is based on the word's surrounding context in the train set and with that pre-defined by the chosen corpus.
Words with a similar semantic meaning should then also have similar vector representations, as measured by their distance in the vector space \citep{heuer_semantic_2015}.
Different methods to pre-compute the embeddings include word2vec \citep{mikolov_efficient_2013}, FastText \citep{bojanowski_enriching_2016} and GloVe \citep{pennington_glove:_2014}.
To make use of the capabilities of pre-trained \acp{lm}, such as BERT \citep{devlin_bert:_2018} or Flair \citep{akbik_contextual_2018}, we evaluate how well their semantic representations perform, by using contextualized word embeddings.
Those are, in contrast to previously mentioned methods, specific to the context of the word in the input sequence.
One major benefit is the fact that the time-consuming feature engineering could become obsolete since the features are implicitly encoded in the word embeddings.
Furthermore, a better semantic representation of the input could lead to better generalization capabilities of the model and, therefore, to better cross-domain performance.\\
This paper answers the following research questions, which will help to assess the importance of the attention layers and contextualized word embeddings for the argument unit segmentation task:
\begin{itemize}
    \item \textbf{RQ1:} To what extent can additional attention layers help the model focus on the, for the task of unit segmentation relevant, sequence parts and how much do they influence the predictions?
    \item \textbf{RQ2:} What is the impact of contextualized distributed term representations like BERT \citep{devlin_bert:_2018} and Flair \citep{akbik_contextual_2018} on the task of unit segmentation and do they improve upon pre-defined representations like GloVe?
\end{itemize}
The contributions of this paper are as follows: first, we present and evaluate new attention based architectures for the task of argumentative text segmentation.
Second, we review the effectiveness of recently proposed contextualized word embedding approaches in regard to \ac{am}.
We will continue by presenting the previous work on this specific task, followed by a description of the data set, the different architectures used and the generation of the word embeddings. Afterward, we will report the results, followed by a discussion and the limitations.
We will finish with a conclusion and an outlook on possible future work.

\section{Related Work}
\label{sec:related-work}
Attention mechanisms have long been utilized in deep neural networks.
Some of its roots are in the salient region detection for the processing of images \citep{itti_model_1998}, which takes inspiration from human perception.
The main idea is to focus the attention of the underlying network on points-of-interest in the input that are often surrounded by irrelevant parts \citep{mnih_recurrent_2014}.
This allows the model to put more weight on the important chunks.
While earlier salient detectors were task-specific, newer approaches (e.g. \citealp{mnih_recurrent_2014}) can be adapted to different tasks, like image description generation \citep{xu_show_2015}, and allow for the parameters of the attention to be tuned during the training.
These additional tasks include sequence processing and the application of such networks to different areas of \ac{nlp}.
One of the first use-cases for attention mechanisms in the field of \ac{nlp} was machine translation.
\citet{bahdanau_neural_2014} utilized the attention to improve their \ac{nmt} model.
A few years later, \citet{vaswani_attention_2017} achieved new \ac{sota} results by presenting an encoder-decoder architecture that is based on the attention mechanism, only adding a position-wise feed-forward network and normalizations in between.
\citet{devlin_bert:_2018} picked up on the encoder part of this architecture to pre-train a bidirectional \ac{lm}.
After fine-tuning, they achieved, again, a new \ac{sota} performance on different downstream \ac{nlp} tasks like Part-of-speech tagging and Questions-Answering.\\
A possible way of posing the unit segmentation as \ac{nlp} task is a token-based sequence labeling \citep{stab_argumentative_2017}.
While \citet{tobias_argument_2018} used rather simple, non-recurrent classifiers to approach this problem, others mostly applied recurrent networks to the task of unit boundary prediction.
For example, \citet{eger_neural_2017} reported different \ac{lstm} \citep{hochreiter_long_1997} architectures.
Further, \citet{ajjour_unit_2017} proposed a setup with three \acp{blstm} \citep{schuster_bidirectional_1997} in total as their best solution.
While the first two of them are fully connected and work on word embeddings and task-specific features respectively, the intention for the third is to take the output of the first two as input and learn to correct their errors.
Even though the third \ac{blstm} did not improve on the F1-score metric, it did succeed in resolving some of the wrong consecutive token predictions, without worsening the final results.\\
To the best of the authors' knowledge, the attention mechanism has not been widely utilized so far for the task of argumentative unit segmentation.
\citet{stab_cross-topic_2018} integrated the attention mechanism directly into their \ac{blstm} by calculating it at each time step $t$ to evaluate the importance of the current hidden state $h_t$.
To do that, they employed additive attention.
A similar approach has been applied by \citet{morio_end--end_2018} for a three-label classification task (claim, premise or non-argumentative).\\
In contrast to that, the approach presented in this paper uses attention as a separate layer that encodes all sequences before they are fed into a \ac{blstm}.
This might enable the recurrent part of the network to learn from better representations that are specific to the task it is trained on.
The aim is further to evaluate the possible applications of attention layers for the task of sequence segmentation and token classification.
A recurrent architecture \citep{ajjour_unit_2017} is compared to multiple modified versions that utilize the attention mechanism.\\
In order to derive a representation of the input text that better resembles the context of the input for a specific task, several approaches have been presented.
\citet{akbik_contextual_2018}, for example, pre-train a character-level \ac{blstm} to predict the next character for a given text corpus.
The pre-trained model is able to derive contextualized word embeddings by additionally utilizing the input sequence for a specific task.
This allows it to encode the previous as well as the following words of the given input sequence into the word itself.
In comparison to that, the pre-trained BERT-\ac{lm} utilizes stacked attention layers \citep{vaswani_attention_2017}.
By feeding a sequence into it and extracting the output of the last sublayer for each token, the idea is to implicitly use the attention mechanism to derive a better representation for every token.
As is the case for the \ac{lm} from \citet{akbik_contextual_2018}, the BERT embeddings are contextualized by the whole input sequence of the specific task.\\
This paper will compare the two contextualized approaches described above with the pre-defined GloVe \citep{pennington_glove:_2014} embeddings in the light of their usefulness for \ac{am}.
The goal is to encode the features necessary to detect arguments by utilizing the context of a sentence.

\section{Methodology}
This paper evaluates different machine learning architectures with added attention layers for the task of \ac{am}, and more specifically unit segmentation. The problem is framed as a multi-class token labeling task, in which each token is assigned one of three labels.
A (B) label denotes that the token is at the beginning of an argumentative unit, an (I) label that it lies inside a unit and an (O) label that the token is not part of a unit.
This framework has been applied previously for the same task \citep{stab_argumentative_2017,eger_neural_2017,ajjour_unit_2017}.\\
The architectures proposed in this section build on \citet{ajjour_unit_2017}, omitting the second \ac{blstm}, which was used to process features other than word embeddings (see section \ref{sec:features}).
They are further being modified by adding attention layers at different positions.
The goal is to reuse existing approaches and possibly enhance their ability to model long-range dependencies.
Additionally, a simpler architecture, consisting of a single \ac{blstm} paired with an attention layer, is built and evaluated with the aim of decreased complexity.\\
In order to answer the second research question, this paper reports results in combination with improved input embeddings, in order to evaluate their effectiveness and impact on the \ac{am} downstream task.\\
All models are compared to the modified re-implementation of the architecture, which is defined as the baseline architecture.

\begin{figure*}[t!]
        \label{fig:architectures}
        \centering
        \begin{subfigure}[t]{0.32\textwidth}
            \centering
            \includegraphics{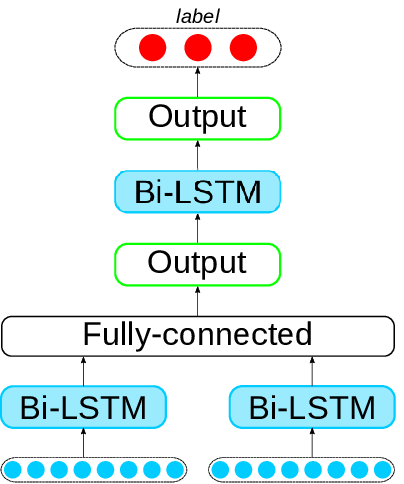}
            \caption{\label{fig:architectures:baseline}}
        \end{subfigure}
        ~
        \begin{subfigure}[t]{0.32\textwidth}
            \centering
            \includegraphics{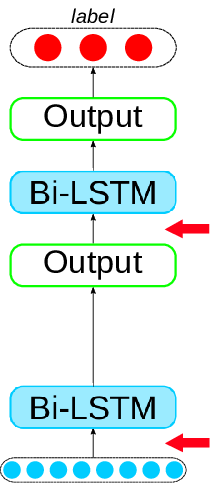}
            \caption{\label{fig:architectures:baseline2}}
        \end{subfigure}
        ~
        \begin{subfigure}[t]{0.32\textwidth}
            \centering
            \includegraphics{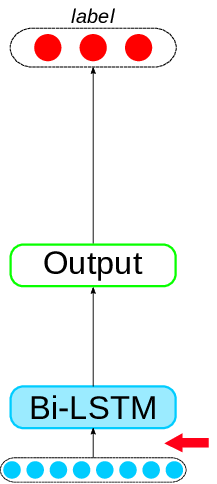}
            \caption{\label{fig:architectures:input-bilstm}}
        \end{subfigure}
        \caption{
            \textit{(\subref{fig:architectures:baseline})} The original baseline architecture as reported by \citet{ajjour_unit_2017}.
            \textit{(\subref{fig:architectures:baseline2})} The modified baseline architecture without the second input Bi-LSTM. The bold arrows show the positions at which the additional attention layers are added to build the \acs{baseline-i} and \acs{baseline-e} architectures.
            \textit{(\subref{fig:architectures:input-bilstm})} The \acs{bilstm} architecture incorporates only one \ac{blstm}. The bold arrow shows the position at which the additional attention layer is added to build the \acs{bilstm-i} architecture.}
    \end{figure*}

\subsection{Features}
\label{sec:features}
    For each token, a set of three different embeddings is generated and compared regarding their capability as standalone input features.
    The resulting weighted F1-score is then used as a proxy for measuring the usefulness of the generated text-representation in light of this specific downstream task.\\
    In combination with the re-implemented architecture, the word vectorization approach GloVe \citep{pennington_glove:_2014}, trained on 6 billion tokens\footnote{\url{https://nlp.stanford.edu/projects/glove/}, last accessed: 2019-04-15, 09:50UTC+2.}, serves as the baseline.\\
    As a first approach to enhance the performance, the GloVe embeddings are stacked with the character-based Flair embeddings \citep{akbik_contextual_2018}, which are generated by a \ac{blstm} model.
    \citet{akbik_contextual_2018} argue that the resulting embeddings are contextualized, since the \ac{lm} was trained to predict the most probable next character and therefore to encode the context of the whole sequence.\\
    Similar to that, we also compare contextualized BERT-embeddings as standalone features \citep{devlin_bert:_2018}.
        An increased performance is expected because of the pre-training procedure of the \ac{lm}.
    The BERT-\ac{lm} was trained to predict a (randomly masked) word by utilizing the context of its appearance, as well as on next sentence prediction.
    Due to its \ac{sota} performance for both, token-level and sentence-level tasks, the authors of this paper argue that the derived representations are well suited for the task of unit segmentation.
    Also, the representation fits the needs of the inter-token and sentence dependencies of the task.
    It is expected that this enables the model to better grasp the notion or pattern of an argument.
    Both contextualized embeddings are generated using the Flair library \citep{zalando_research_very_2018}.\\
    Features specifically engineered for this task are not included in the input, following the argumentation of \citet{eger_neural_2017} that they will probably not be generalizable to different data sets.

\subsection{Data}
    In order to evaluate the different architectures, the ``Argument annotated Essays (version 2)'' corpus (also referred to as Persuasive Essays corpus) is used \citep{stab_parsing_2017}.
    It was utilized for the same task in previous literature \citep{ajjour_unit_2017,eger_neural_2017}.\\
    The corpus, compiled for parsing argumentative structures in written text, consists of a random sample of 402 student essays.
    The annotation scheme includes the argumentative units and the relations between them, as well as the major claim and stance of the author towards a specific topic.
    The texts were annotated by non-professionals, labeling the boundary of each argumentative unit alongside the unit type.
    A type can either be major-claim, claim or premise.
    For the unit segmentation task, the corpus is labeled by treating major claims, claims, and premises as argumentative units\footnote{All data pre-processing scripts are available in our code repository: \url{https://gitlab.informatik.uni-bremen.de/covis1819/worth-the-attention}.}.
    For comparability reasons in the evaluation process, the models are trained and tested with the train-test-split defined by \citet{stab_parsing_2017}.

\subsection{Models}
\label{sec:models}
    In order to evaluate the attention mechanisms, different architectures based on previous \ac{am} literature are implemented.
    The attention layer is added at different positions in the network.\\
    All models were implemented using Python and the Keras framework with a TensorFlow backend.
    For the self-attention and multi-head attention layers, an existing implementation is used \citep{hg_attention_2018,hg_wrapper_2018}.
    The difference between the two is that the multi-head attention divides the input into multiple chunks and each head therefore works on a different vector subspace \citep{vaswani_attention_2017}, while the self-attention works on the whole input sequence.
    This is supposed to allow the head to focus on specific features of the input.
    In this case, the self-attention layers use additive attention, while the multi-head attention layers use scaled dot-product attention, with the latter following the implementation of \citet{vaswani_attention_2017}.

    \paragraph*{Baseline re-implementation}
    The baseline model from \citet{ajjour_unit_2017} uses a total of three \acp{blstm} (two of them fully connected) to assign labels to tokens (see Figure~\ref{fig:architectures:baseline}).
    The re-implementation does not include the two fully connected \acp{blstm} but instead uses only a single one that works on the word embeddings (see Figure~\ref{fig:architectures:baseline2}).
    Due to the fact that the second \ac{blstm} in the first layer is only used to encode the non-semantic features like Part-of-speech tags and discourse marker labels, it is omitted in the re-implementation.
    Hereafter, we will refer to this model as \acs{baseline}.
    Also, the batch size was increased from 8 to 64, compared to the original implementation, as a trade-off between convergence time and the model's generalization performance \citep{keskar_large-batch_2016}.
    Nevertheless, this model achieves comparable scores to the ones presented in the original paper.
    The slightly lower performance can probably be attributed to implementation details.

    \paragraph*{\acs{baseline-i} and \acs{baseline-e}}
    For both variations, the baseline architecture was used as a basis, as can be seen in Figure~\ref{fig:architectures:baseline2}.
    Multi-head-attention layers are added at different positions in the network.
    The number of attention heads depends on the dimension of the embedding vectors.
    For the GloVe (300 features) and the BERT (3072 features) embeddings, six heads are used, while the Flair (4196 features) embeddings require four heads.
    Both numbers were the largest divisor for the respective input vector size that worked inside the computational boundaries available.
    In the first model, an attention layer was added before the first \ac{blstm} in an attempt to apply a relevance score directly to the tokens, in order to better capture dependencies of the input sequence.
    This model will be referred to as \acs{baseline-i}.
    The second variation adds the attention layer after the first and before the second \ac{blstm}, which will be called \acs{baseline-e}.
    According to \citet{ajjour_unit_2017}, the latter \ac{blstm} is used to correct the errors of the first one.
    The attention layer should be able to support the model in the error correction process.
    In contrast to the first approach, this does not change the input data, but only works on the output of the first \ac{blstm}.

    \paragraph*{\acs{bilstm} and \acs{bilstm-i}}
    To decrease the complexity of the architecture, two additional models with a single \ac{blstm} are trained.
    The first variant has no attention layer, while the second one utilized the same input attention described above (see Figure~\ref{fig:architectures:input-bilstm}).
    They will be refered to as \acs{bilstm} and \acs{bilstm-i} respectively.
    Both architectures use a self-attention mechanism instead of the above-mentioned multi-head-attention, due to better results in preliminary tests.

\section{Results}
\label{sec:results}

\newcommand\innerspacing{\rule[-2.0ex]{0pt}{0pt}} 
\begin{table}[t!]
    \centering
    \begin{tabular}{lccc}
        \specialrule{.1em}{.05em}{.05em}
        Model \rule{0pt}{3.0ex}        & GloVe          & BERT          & Flair         \\ \hline
        \acs{baseline} \innerspacing   & 0.86           & 0.83          & \textbf{0.87} \\
        \acs{baseline-i}               & \textbf{0.85}  & 0.68          & 0.67          \\
        \acs{baseline-e} \innerspacing & 0.67           & \textbf{0.68} & 0.67          \\
        \acs{bilstm}                   & \textbf{0.86}  & 0.86          & 0.86          \\
        \acs{bilstm-i} \innerspacing   & \textbf{0.84}  & 0.83          & 0.81          \\
        \specialrule{.1em}{.05em}{.05em}
    \end{tabular}
    \caption{The weighted F1-scores for the \acs{baseline} and all four variations. Results are shown per variation and embedding. Each row shows the performance of one architecture with different word embeddings as input vector. The highest score for each architecture is marked in bold.}
    \label{tab:results}
\end{table}

We evaluate the performance of all architectures on the Persuasive Essays data set detailed above.
The models are re-initialized after every evaluation and do not share any weights.
This allows us to answer the first research question of whether additional attention layers have a positive impact on the prediction quality.\\
To answer the second research question, we re-run each training, replacing the GloVe with BERT and Flair embeddings.
Both contextualized embedding methods are tested separately.
We contextualize the tokens on the sentence level since the BERT model \citep{google_research_tensorflow_2018} only allows for a maximum input length of 512 characters.
This makes document-level or paragraph-level embeddings impractical for the data set.\\
As a performance measure, we report the weighted F1-score instead of the macro F1-score, since it takes the imbalance of the samples per label into account.\\
For our re-implementation of the baseline, we are able to approximately reproduce the results reported by \citet{ajjour_unit_2017}.
Additionally, we can verify that there is no major change in the performance when adding a second \ac{blstm} to the network (compare results for \acs{bilstm} and \acs{baseline} in Table \ref{tab:results}).

\subsection{Attention Layers}
    The results of the token classification task are presented in Table \ref{tab:results}.
    Generally speaking, the added attention encodings do not improve upon the original architecture's performance, no matter at which position they are added.
    Architectures with an input attention encoding, namely \acs{baseline-i} and \acs{bilstm-i}, do achieve similar performances compared to their respective baseline.
    But the F1-score performance is in strong contrast to the generalization error, which is in most cases lower for the \acs{baseline} model.\\
    The \acs{baseline-e} architecture on the other hand, which is supposed to help the second \ac{blstm} in the network to correct the errors made by the first one, performs worse across all tests.
    For the Flair embeddings, this results in a 0.20 points performance drop in the F1-score measure.

\begin{figure*}[t!]
    \centering
    \begin{subfigure}[t]{0.32\textwidth}
        \centering
        \includegraphics[scale=0.35]{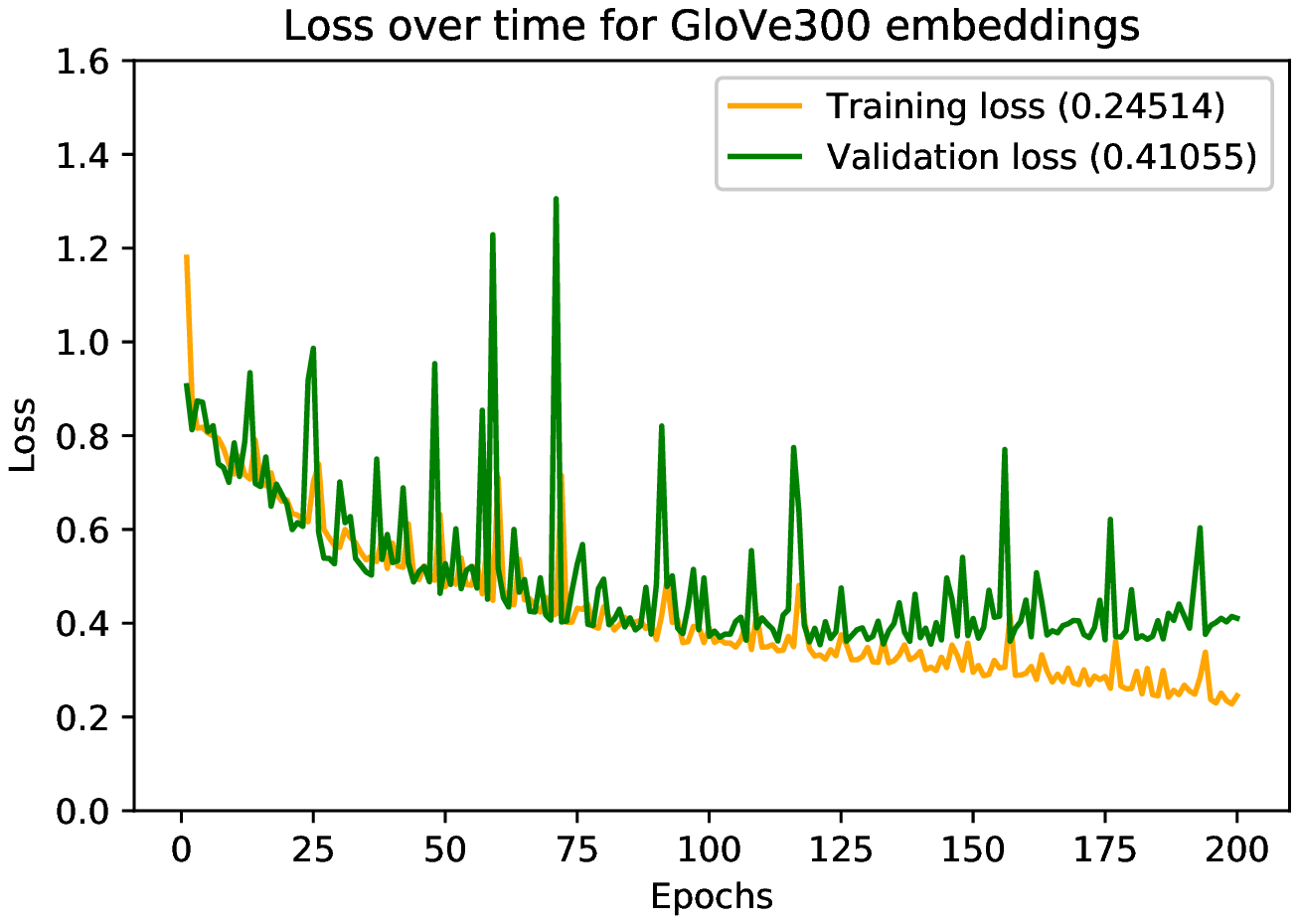}
        \caption{\label{fig:loss:glove}}
    \end{subfigure}
    ~
    \begin{subfigure}[t]{0.32\textwidth}
        \centering
        \includegraphics[scale=0.35]{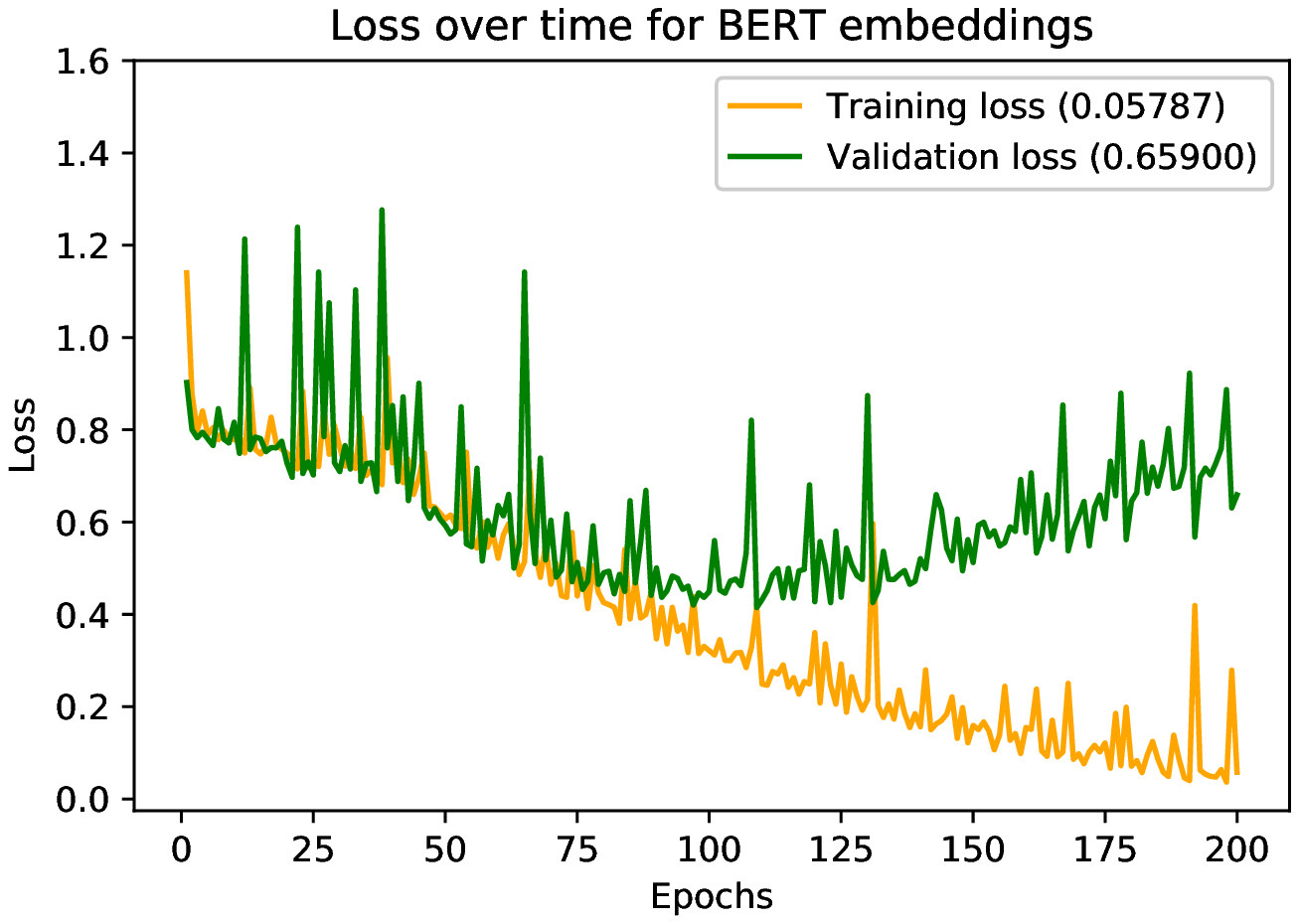}
        \caption{\label{fig:loss:bert}}
    \end{subfigure}
    ~
    \begin{subfigure}[t]{0.32\textwidth}
        \centering
        \includegraphics[scale=0.35]{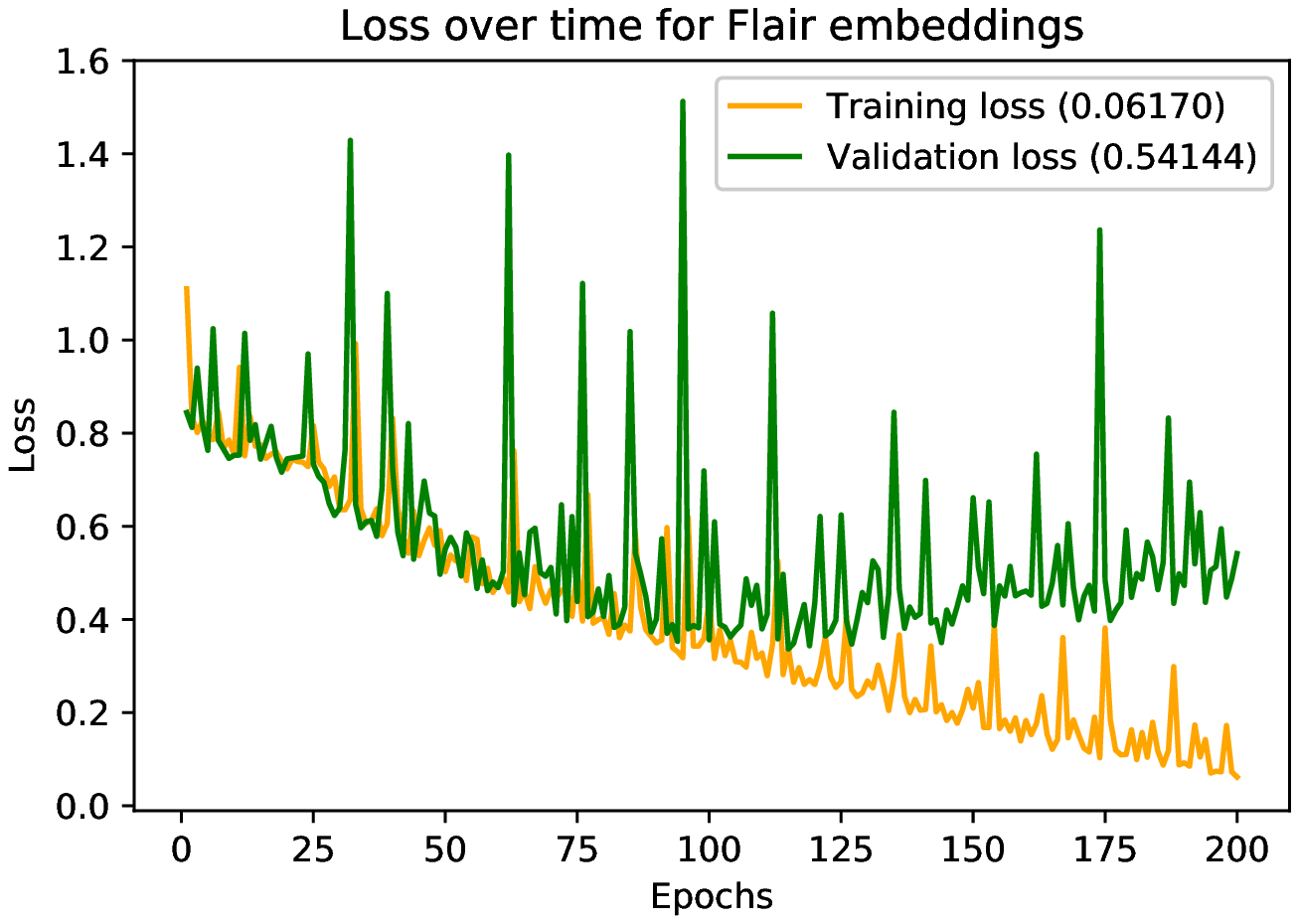}
        \caption{\label{fig:loss:flair}}
    \end{subfigure}
    \caption{The loss curves of the \acs{baseline} architecture using different input embeddings. \textit{(\subref{fig:loss:glove})} shows the training process of the model using the GloVe embeddings, while the model in \textit{(\subref{fig:loss:bert})} used the BERT embeddings and \textit{(\subref{fig:loss:flair})} the Flair embeddings. The bottom orange line shows the training loss, the top green line the validation loss.}
    \label{fig:loss}
\end{figure*}

\subsection{Contextualized Word Embeddings}
    The results for the enhanced word embedding evaluations are reported in Table \ref{tab:results}.
    In some cases, the models utilizing the word embeddings generated by the BERT-\ac{lm} achieve a lower performance score than the other embeddings.
    This drop is most noticeable for the \acs{baseline-i} model, while the performance for the \acs{bilstm-i} decreases only slightly.
    The \acs{baseline-e} model is able to achieve results that outperform both, GloVe and Flair embeddings.\\
    Compared to the GloVe vectors, the models trained on the Flair embeddings mostly lose in F1-score performance as well.
    For example, the \acs{baseline-i} model drops by 0.18 points.
    On the other hand, the \acs{baseline} model is able to slightly improve upon the GloVe score using the Flair embeddings, achieving a final score of 0.87, which also marks the best overall score in our testings.\\
    An interesting observation is the fact that the enhanced embeddings seem to increase the generalization error (compare Figure \ref{fig:loss}).
    The \acs{baseline} model trained on the GloVe embeddings for example, shows a difference in the final validation and training loss of around 0.17 and increases for the BERT and Flair embeddings to roughly 0.60 and 0.48 points, respectively.

\section{Discussion}
Given the experimental results, we discuss the resulting implications for our two research questions and conclude this section by presenting some limitations.
\subsection{Attention Layers}
    Our results suggest that the attention encoding does not increase the performance of the model, as we hypothesized above.
    This is true for both, the input and the error encoding.
    A potential explanation is the fact that we use the attention mechanism as an additional layer to encode the input.
    Other approaches, like \citet{morio_end--end_2018} or \citet{stab_cross-topic_2018}, incorporate it into the \ac{blstm} architecture and calculate the weight of the hidden states at every time step.\\
    While the performance does not decrease meaningfully for the \acs{baseline-i} and \acs{bilstm-i} models (using the GloVe embeddings as features), it does for the error encoding \acs{baseline-e} model.
    This drop might be explained by the vector space the attention mechanism is working on.
    Due to its small size of only four features, it is unlikely that the resulting vector has a meaningful encoding.\\
    A deeper inspection of the output values from the different layers in the network and how they influence the overall classification task might give more insight into the cause of the problem.

\subsection{Contextualized Word Embeddings}
    For most of the tests we conduct, the contextualized embedding approaches do not improve upon the GloVe embeddings.
    This is especially true for the architectures that include an attention layer, which does not seem to be able to handle the encoding of high dimensional vectors very well.
    The results further suggest that the amount of neurons in the \acp{blstm} is not an issue in this case, since the \acs{baseline} model achieves comparable results across all three embeddings.\\
    A potential way to improve the results of the enhanced embeddings is to contextualize them on the paragraph level.
    While we contextualize them on a sentence level, the dependencies between arguments might span over multiple sentences, sometimes even a paragraph, as described by \citet{stab_parsing_2017} for the Persuasive Essays data set.
    Following this reasoning, one might think that a document level contextualization makes sense and adds even more information to the embedding.
    For the task of AM, however, we argue against that for two reasons.
    First, argumentative units usually do not span over the whole document and it might include additional counter-arguments \citep{stab_parsing_2017}.
    The contextualization would most likely cause a lot of noise and make the vector less useful.
    Also, depending on the size of the document, the size of the vector might be too small to hold the contextual information of the full document.
    Second, the model trained on such embeddings would probably not generalize very well.
    An argumentative document can be written in different formats with different purposes, like an essay, a speech or a newspaper article.
    Contextualizing the embeddings on the document level might then also encode the structure of the text and decrease the cross-domain applicability of the model.

\subsection{Limitations}
\label{sec:limitations}
    The results we report and analyze above are the networks' performance as validated on the data splits provided by \citet{stab_parsing_2017}.
    Due to time and resource restrictions, we evaluate the results after a single training run and perform neither an averaging over multiple runs nor any cross-validation.
    Both could lead to more reliable results.
    As another consequence of the above-mentioned restrictions, we are also not able to test the model's generalization capabilities on different data sets.\\
    For the learning rate, we perform only a basic Bayesian hyperparameter optimization \citep{snoek_practical_2012} with four iterations per model.
    These limitations are especially important for the variations of the \acs{baseline} architecture, since the performed changes to the architecture, even though rather small, entail the need for independently tuned hyperparameters.\\
    Furthermore, an additional evaluation of the different contextualization levels for the embeddings could provide a clearer picture of how much the results actually improve, compared to non-contextualized methods.

\section{Conclusion}
Recent improvements in utilizing contextual information for sequence processing had a big impact on the area of \ac{nlp}, namely advances of attention architectures and contextualized word embeddings.
For example, the Transformer architecture \citep{vaswani_attention_2017} employs attention to achieve \ac{sota} scores on different \ac{nlp} tasks.
Further, the Flair model \citep{akbik_contextual_2018} incorporates character-wise context to generate enhanced word representations.\\
In this paper, we report on the usefulness of these two approaches for the task of \ac{am}.
First, we are able to show that an attention layer as additional encoding of the input does not improve upon the current \ac{sota} approach of a \ac{blstm}.
Additionally, the attention mechanism seems to fail for a low-dimensional vector space.
Second, we present the impact of contextualized word embeddings for \ac{am}.
Although the Flair embeddings slightly improve upon the performance of the GloVe embeddings for the \acs{baseline} architecture, we can not confirm any advantage over non-contextualized embeddings.

\subsection{Future Work}
    A first extension of this work could be a proper hyperparameter optimization for the attention-based models.
    Second, we plan to explore an attempt to fine-tune solely attention based pre-trained models like BERT \citep{devlin_bert:_2018} to domain-specific data.
    Recent research by \citet{howard_universal_2018} in transfer-learning for \ac{nlp} has shown great improvement for several \ac{nlp}-downstream tasks, while reducing the needed amount of labeled training data.\\
    Third, we contextualize the embeddings on the sentence level only.
    According to \citet{stab_parsing_2017}, arguments can sometimes span over multiple sentences.
    Therefore, the contextualization of the embeddings could be extended to a paragraph level, in order to make use of possible inter-dependencies within it.
    Additionally, a fine-tuning approach of the underlying \ac{lm}s to the \ac{am} task could further enhance the embeddings.


\bibliography{acl2019}

\begin{thebibliography}{28}
\expandafter\ifx\csname natexlab\endcsname\relax\def\natexlab#1{#1}\fi

\bibitem[{Ajjour et~al.(2017)Ajjour, Chen, Kiesel, Wachsmuth, and
  Stein}]{ajjour_unit_2017}
Yamen Ajjour, Wei-Fan Chen, Johannes Kiesel, Henning Wachsmuth, and Benno
  Stein. 2017.
\newblock \href {https://doi.org/10.18653/v1/W17-5115} {Unit {Segmentation} of
  {Argumentative} {Texts}}.
\newblock In \emph{Proceedings of the 4th {Workshop} on {Argument} {Mining}},
  pages 118--128, Copenhagen, Denmark. Association for Computational
  Linguistics.

\bibitem[{Akbik et~al.(2018)Akbik, Blythe, and
  Vollgraf}]{akbik_contextual_2018}
Alan Akbik, Duncan Blythe, and Roland Vollgraf. 2018.
\newblock \href {http://aclweb.org/anthology/C18-1139} {Contextual {String}
  {Embeddings} for {Sequence} {Labeling}}.
\newblock In \emph{Proceedings of the 27th {International} {Conference} on
  {Computational} {Linguistics}}, pages 1638--1649, Santa Fe, New Mexico, USA.
  Association for Computational Linguistics.

\bibitem[{Bahdanau et~al.(2014)Bahdanau, Cho, and
  Bengio}]{bahdanau_neural_2014}
Dzmitry Bahdanau, Kyunghyun Cho, and Yoshua Bengio. 2014.
\newblock \href {http://arxiv.org/abs/1409.0473} {Neural {Machine}
  {Translation} by {Jointly} {Learning} to {Align} and {Translate}}.
\newblock arXiv: 1409.0473.

\bibitem[{Bojanowski et~al.(2017)Bojanowski, Grave, Joulin, and
  Mikolov}]{bojanowski_enriching_2016}
Piotr Bojanowski, Edouard Grave, Armand Joulin, and Tomas Mikolov. 2017.
\newblock \href {https://doi.org/10.1162/tacl_a_00051} {Enriching {Word}
  {Vectors} with {Subword} {Information}}.
\newblock \emph{Transactions of the Association for Computational Linguistics},
  5:135--146.

\bibitem[{Cabrio and Villata(2018)}]{cabrio_five_2018}
Elena Cabrio and Serena Villata. 2018.
\newblock \href {https://doi.org/10.24963/ijcai.2018/766} {Five {Years} of
  {Argument} {Mining}: a {Data}-driven {Analysis}}.
\newblock In \emph{Proceedings of the {Twenty}-{Seventh} {International}
  {Joint} {Conference} on {Artificial} {Intelligence}}, pages 5427--5433,
  Stockholm, Sweden. International Joint Conferences on Artificial Intelligence
  Organization.

\bibitem[{Devlin et~al.(2018)Devlin, Chang, Lee, and
  Toutanova}]{devlin_bert:_2018}
Jacob Devlin, Ming-Wei Chang, Kenton Lee, and Kristina Toutanova. 2018.
\newblock \href {http://arxiv.org/abs/1810.04805} {{BERT}: {Pre}-training of
  {Deep} {Bidirectional} {Transformers} for {Language} {Understanding}}.
\newblock arXiv: 1810.04805.

\bibitem[{Eger et~al.(2017)Eger, Daxenberger, and Gurevych}]{eger_neural_2017}
Steffen Eger, Johannes Daxenberger, and Iryna Gurevych. 2017.
\newblock \href {https://doi.org/10.18653/v1/P17-1002} {Neural {End}-to-{End}
  {Learning} for {Computational} {Argumentation} {Mining}}.
\newblock In \emph{Proceedings of the 55th {Annual} {Meeting} of the
  {Association} for {Computational} {Linguistics} ({Volume} 1: {Long}
  {Papers})}, pages 11--22, Vancouver, Canada. Association for Computational
  Linguistics.

\bibitem[{{Google AI Research}(2018)}]{google_research_tensorflow_2018}
{Google AI Research}. 2018.
\newblock \href {https://github.com/google-research/bert} {{TensorFlow} code
  and pre-trained models for {BERT}.}
\newblock \url{https://github.com/google-research/bert}, last accessed:
  2019-05-01, 21:40UTC+2.

\bibitem[{Heuer(2015)}]{heuer_semantic_2015}
Hendrik Heuer. 2015.
\newblock \href {https://aaltodoc.aalto.fi:443/handle/123456789/17732}
  {Semantic and stylistic text analysis and text summary evaluation}.
\newblock Master thesis.

\bibitem[{HG(2018{\natexlab{a}})}]{hg_attention_2018}
Zhao HG. 2018{\natexlab{a}}.
\newblock \href {https://github.com/CyberZHG/keras-self-attention} {Attention
  mechanism for processing sequential data that considers the context for each
  timestamp.}
\newblock \url{https://github.com/CyberZHG/keras-self-attention}, last
  accessed: 2019-05-01, 21:39UTC+2.

\bibitem[{HG(2018{\natexlab{b}})}]{hg_wrapper_2018}
Zhao HG. 2018{\natexlab{b}}.
\newblock \href {https://github.com/CyberZHG/keras-multi-head} {A wrapper layer
  for stacking layers horizontally.}
\newblock \url{https://github.com/CyberZHG/keras-multi-head}, last accessed:
  2019-05-01, 21:40UTC+2.

\bibitem[{Hochreiter and Schmidhuber(1997)}]{hochreiter_long_1997}
Sepp Hochreiter and Jürgen Schmidhuber. 1997.
\newblock \href {https://doi.org/10.1162/neco.1997.9.8.1735} {Long
  {Short}-{Term} {Memory}}.
\newblock \emph{Neural Computation}, 9(8):1735--1780.

\bibitem[{Howard and Ruder(2018)}]{howard_universal_2018}
Jeremy Howard and Sebastian Ruder. 2018.
\newblock \href {https://aclweb.org/anthology/papers/P/P18/P18-1031/}
  {Universal {Language} {Model} {Fine}-tuning for {Text} {Classification}}.
\newblock In \emph{Proceedings of the 56th {Annual} {Meeting} of the
  {Association} for {Computational} {Linguistics} ({Volume} 1: {Long}
  {Papers})}, pages 328--339, Melbourne, Australia. Association for
  Computational Linguistics.

\bibitem[{Itti et~al.(1998)Itti, Koch, and Niebur}]{itti_model_1998}
Laurent Itti, Christof Koch, and Ernst Niebur. 1998.
\newblock \href {https://doi.org/10.1109/34.730558} {A model of saliency-based
  visual attention for rapid scene analysis}.
\newblock \emph{IEEE Transactions on Pattern Analysis and Machine
  Intelligence}, 20(11):1254--1259.

\bibitem[{Keskar et~al.(2016)Keskar, Mudigere, Nocedal, Smelyanskiy, and
  Tang}]{keskar_large-batch_2016}
Nitish~Shirish Keskar, Dheevatsa Mudigere, Jorge Nocedal, Mikhail Smelyanskiy,
  and Ping Tak~Peter Tang. 2016.
\newblock \href {http://arxiv.org/abs/1609.04836} {On {Large}-{Batch}
  {Training} for {Deep} {Learning}: {Generalization} {Gap} and {Sharp}
  {Minima}}.
\newblock arXiv: 1609.04836.

\bibitem[{Mikolov et~al.(2013)Mikolov, Chen, Corrado, and
  Dean}]{mikolov_efficient_2013}
Tomas Mikolov, Kai Chen, Greg Corrado, and Jeffrey Dean. 2013.
\newblock \href {http://arxiv.org/abs/1301.3781} {Efficient {Estimation} of
  {Word} {Representations} in {Vector} {Space}}.
\newblock arXiv: 1301.3781.

\bibitem[{Mnih et~al.(2014)Mnih, Heess, Graves, and
  Kavukcuoglu}]{mnih_recurrent_2014}
Volodymyr Mnih, Nicolas Heess, Alex Graves, and Koray Kavukcuoglu. 2014.
\newblock \href {http://arxiv.org/abs/1406.6247} {Recurrent {Models} of
  {Visual} {Attention}}.
\newblock arXiv: 1406.6247.

\bibitem[{Morio and Fujita(2018)}]{morio_end--end_2018}
Gaku Morio and Katsuhide Fujita. 2018.
\newblock \href {https://aclweb.org/anthology/papers/W/W18/W18-5202/}
  {End-to-{End} {Argument} {Mining} for {Discussion} {Threads} {Based} on
  {Parallel} {Constrained} {Pointer} {Architecture}}.
\newblock In \emph{Proceedings of the 5th {Workshop} on {Argument} {Mining}},
  pages 11--21, Brussels, Belgium. Association for Computational Linguistics.

\bibitem[{Pennington et~al.(2014)Pennington, Socher, and
  Manning}]{pennington_glove:_2014}
Jeffrey Pennington, Richard Socher, and Christopher Manning. 2014.
\newblock \href {https://doi.org/10.3115/v1/D14-1162} {Glove: {Global}
  {Vectors} for {Word} {Representation}}.
\newblock In \emph{Proceedings of the 2014 {Conference} on {Empirical}
  {Methods} in {Natural} {Language} {Processing} ({EMNLP})}, pages 1532--1543,
  Doha, Qatar. Association for Computational Linguistics.

\bibitem[{Schuster and Paliwal(1997)}]{schuster_bidirectional_1997}
Mike Schuster and Kuldip~K. Paliwal. 1997.
\newblock \href {https://doi.org/10.1109/78.650093} {Bidirectional recurrent
  neural networks}.
\newblock \emph{IEEE Trans. Signal Processing}, 45:2673--2681.

\bibitem[{Snoek et~al.(2012)Snoek, Larochelle, and
  Adams}]{snoek_practical_2012}
Jasper Snoek, Hugo Larochelle, and Ryan~P Adams. 2012.
\newblock \href
  {http://papers.nips.cc/paper/4522-practical-bayesian-optimization-of-machine-learning-algorithms.pdf}
  {Practical {Bayesian} {Optimization} of {Machine} {Learning} {Algorithms}}.
\newblock In F.~Pereira, C.~J.~C. Burges, L.~Bottou, and K.~Q. Weinberger,
  editors, \emph{Advances in {Neural} {Information} {Processing} {Systems} 25},
  pages 2951--2959. Curran Associates, Inc.

\bibitem[{Stab and Gurevych(2017)}]{stab_parsing_2017}
Christian Stab and Iryna Gurevych. 2017.
\newblock \href {https://doi.org/10.1162/COLI_a_00295} {Parsing {Argumentation}
  {Structures} in {Persuasive} {Essays}}.
\newblock \emph{Computational Linguistics}, 43(3):619--659.

\bibitem[{Stab et~al.(2018)Stab, Miller, Schiller, Rai, and
  Gurevych}]{stab_cross-topic_2018}
Christian Stab, Tristan Miller, Benjamin Schiller, Pranav Rai, and Iryna
  Gurevych. 2018.
\newblock \href {http://aclweb.org/anthology/D18-1402} {Cross-topic {Argument}
  {Mining} from {Heterogeneous} {Sources}}.
\newblock In \emph{Proceedings of the 2018 {Conference} on {Empirical}
  {Methods} in {Natural} {Language} {Processing}}, pages 3664--3674.
  Association for Computational Linguistics.
\newblock Event-place: Brussels, Belgium.

\bibitem[{Stab(2017)}]{stab_argumentative_2017}
Christian Matthias~Edwin Stab. 2017.
\newblock \href {http://tuprints.ulb.tu-darmstadt.de/6006/}
  {\emph{Argumentative {Writing} {Support} by means of {Natural} {Language}
  {Processing}}}.
\newblock Dissertation, Technische Universität Darmstadt, Darmstadt.

\bibitem[{Tobias et~al.(2018)Tobias, Elena, Marco, Paolo, and
  Serena}]{tobias_argument_2018}
Mayer Tobias, Cabrio Elena, Lippi Marco, Torroni Paolo, and Villata Serena.
  2018.
\newblock \href {https://doi.org/10.3233/978-1-61499-906-5-137} {Argument
  {Mining} on {Clinical} {Trials}}.
\newblock \emph{Frontiers in Artificial Intelligence and Applications}, pages
  137--148.

\bibitem[{Vaswani et~al.(2017)Vaswani, Shazeer, Parmar, Uszkoreit, Jones,
  Gomez, Kaiser, and Polosukhin}]{vaswani_attention_2017}
Ashish Vaswani, Noam Shazeer, Niki Parmar, Jakob Uszkoreit, Llion Jones,
  Aidan~N. Gomez, Lukasz Kaiser, and Illia Polosukhin. 2017.
\newblock \href {http://dl.acm.org/citation.cfm?id=3295222.3295349} {Attention
  is {All} {You} {Need}}.
\newblock In \emph{Proceedings of the 31st {International} {Conference} on
  {Neural} {Information} {Processing} {Systems}}, {NIPS}'17, pages 6000--6010,
  USA. Curran Associates Inc.
\newblock Event-place: Long Beach, California, USA.

\bibitem[{Xu et~al.(2015)Xu, Ba, Kiros, Cho, Courville, Salakhudinov, Zemel,
  and Bengio}]{xu_show_2015}
Kelvin Xu, Jimmy Ba, Ryan Kiros, Kyunghyun Cho, Aaron Courville, Ruslan
  Salakhudinov, Rich Zemel, and Yoshua Bengio. 2015.
\newblock \href {http://proceedings.mlr.press/v37/xuc15.html} {Show, {Attend}
  and {Tell}: {Neural} {Image} {Caption} {Generation} with {Visual}
  {Attention}}.
\newblock In \emph{Proceedings of the 32nd {International} {Conference} on
  {Machine} {Learning}}, volume~37 of \emph{Proceedings of {Machine} {Learning}
  {Research}}, pages 2048--2057, Lille, France. PMLR.

\bibitem[{{Zalando Research}(2018)}]{zalando_research_very_2018}
{Zalando Research}. 2018.
\newblock \href {https://github.com/zalandoresearch/flair} {A very simple
  framework for state-of-the-art {Natural} {Language} {Processing} ({NLP})}.
\newblock \url{https://github.com/zalandoresearch/flair}, last accessed:
  2019-05-01, 21:39UTC+2.

\end{thebibliography}
\bibliographystyle{acl_natbib}

\end{document}